\pdfoutput=1
\documentclass[conference]{IEEEtran}
\IEEEoverridecommandlockouts
\usepackage{cite}
\usepackage{amsmath,amssymb,amsfonts}
\usepackage{algorithmic}
\usepackage{graphicx}
\usepackage{textcomp}
\usepackage{xcolor}
\usepackage{siunitx}
\usepackage[bookmarks=false]{hyperref}
\usepackage{amssymb}
\usepackage{pgfplots}
\usepackage[acronym,nomain,nonumberlist]{glossaries}
\pgfplotsset{compat=1.14}
\newacronym{ahmpc}{AHMPC}{Adaptive Horizon Model Predictive Control}
\newacronym{admm}{ADMM}{Alternating Direction Method of Multipliers}
\newacronym{ahe}{AHE}{Adaptive Histogram Equalization}
\newacronym{ar}{AR}{Augmented Reality}
\newacronym{clahe}{CLAHE}{Contrast Limited Adaptive Histogram Equalization}
\newacronym{cnn}{CNN}{Convolutional Neural Network}
\newacronym{cog}{CoG}{Center of Gravity}
\newacronym{dbscan}{DBSCAN}{Density-Based Spatial Clustering of Applications with Noise }
\newacronym{dl}{DL}{Deep Learning}
\newacronym{dof}{DoF}{Degree of Freedom}
\newacronym{ekf}{EKF}{Extended Kalman Filter}
\newacronym{fbe}{FBE}{Forward Backward Envelope}
\newacronym{fov}{FoV}{Field of View}
\newacronym{fps}{fps}{Frame Per Second}
\newacronym{gc}{GC}{Geomtric Center}
\newacronym{gnss}{GNSS}{Global Navigation Satellite System}
\newacronym{gps}{GPS}{Global Positioning System}
\newacronym{iid}{IID}{Independent Identically Distributed}
\newacronym{imu}{IMU}{Inertial Measurement Unit}
\newacronym{kf}{KF}{Extended Kalman Filter}
\newacronym{llwcf}{LLWCF}{Left Local Wall Continuum Function}
\newacronym{lpv}{LPV}{Linear Parameter Varying}
\newacronym{lrcn}{LRCN}{ Long-Term Recurrent Convolutional Network}
\newacronym{lwcf}{LWCF}{Local Wall Continuum Function}
\newacronym{mae}{MAE}{mean absolute error}
\newacronym{mav}{MAV}{Micro Aerial Vehicle}
\newacronym{mhe}{MHE}{Moving Horizon Estimation}
\newacronym{mimo}{MIMO}{Multiple Input Multiple Output}
\newacronym{mlp}{MLP}{MultiLayer Perceptron}
\newacronym{moi}{MoI}{Moments of Inertia}
\newacronym{mpc}{MPC}{Model Predictive Control}
\newacronym{msf}{MSF}{Multi Sensor Fusion}
\newacronym{nmhe}{NMHE}{Nonlinear Moving Horizon Estimation}
\newacronym{nmpc}{NMPC}{Nonlinear Model Predictive Control}
\newacronym{nn}{NN}{Nueral Network}
\newacronym{nwu}{NWU}{North-West-Up}
\newacronym{open}{OpEn}{Optimization Engine}
\newacronym{panoc}{PANOC}{Proximal Averaged Newton-type method for Optimal Control}
\newacronym{pdf}{PDF}{Probability Density Function}
\newacronym{pid}{PID}{Proportional Integral Derivative}
\newacronym{psd}{PSD}{ Positive Semi-Definite }
\newacronym{rbf}{RBF}{ Radial Basis Function}
\newacronym{relu}{ReLU}{Rectified Linear Unit}
\newacronym{rl}{RL}{Reinforcement Learning}
\newacronym{rlwcf}{RLWCF}{Right Local Wall Continuum Function}
\newacronym{rmse}{RMSE}{Root Mean Square Error}
\newacronym{ros}{ROS}{Robot Operating System}
\newacronym{sfm}{SfM}{Structure from Motion}
\newacronym{spams}{SPAMS}{SPArse Modeling Software}
\newacronym{sqp}{SQP}{Sequential Quadratic Programming}
\newacronym{ugv}{UGV}{Unmanned Ground Vehicle}
\newacronym{ukf}{UKF}{Unscented Kalman Filter}
\newacronym{vi}{VI}{Visual Inertia}
\newacronym{vo}{VO}{Visual Odometry}

\def\BibTeX{{\rm B\kern-.05em{\sc i\kern-.025em b}\kern-.08em
 T\kern-.1667em\lower.7ex\hbox{E}\kern-.125emX}}
\begin{document}

\title{Switching Model Predictive Control for Online Structural Reformations of a Foldable Quadrotor \\
\thanks{This work has been partially funded by the European Unions Horizon 2020 Research and Innovation Programme Illumineation and Interreg Nord Programme ROBOSOL NYPS 20202891.}
}
\author{\IEEEauthorblockN{Andreas Papadimitriou and George Nikolakopoulos}
\IEEEauthorblockA{
Robotics and AI Team\\ Department of Computer Science, Electrical and Space Engineering\\ Lule\r{a} University of Technology \\ Lule\r{a} SE-97187, Sweden
}
}
\maketitle

\begin{abstract}
The aim of this article is the formulation of a switching model predictive control framework for the case of a foldable quadrotor with the ability to retain the overall control quality during online structural reformations. The majority of the related scientific publications consider fixed morphology of the aerial vehicles. Recent advances in mechatronics have brought novel considerations for generalized aerial robotic designs with the ability to alter their morphology in order to adapt to their environment, thus enhancing their capabilities. Simulation results are provided to prove the efficacy of the selected control scheme.
\end{abstract}

\glsresetall 
\section{Introduction} \label{sec:intro}
Recent advances in technology have made possible the use of aerial vehicles in a wide range of applications ranging from inspection and maintenance \cite{ mansouri2018cooperative,MELO2017174,Kochetkova2018} to exploration \cite{TheShapeshifter,mansouri2020deploying}, search and rescue missions \cite{tomic2012toward,sampedro2018fully}, etc. 

Much research focused to tackle the challenge of fully automated solutions while using fixed-frame quadrotors. To further, increase the variety of tasks and corresponding applications, the ability of a \gls{mav} to alter its structure should be further investigated. One way of achieving structural reformation of the quadrotor is to add \gls{dof} to enable the motion of the quadrotor arms. However, the lack of a generalized control scheme to adapt and capture the dynamics of this frame reconfiguration limits the adaptability of transformable aerial vehicles to different flight conditions.
\subsection{Background \& Motivation}

In the related literature, can be found some aerial vehicles which have the ability to alter their structural formation. The starting point for this article has been the development of a foldable quadrotor with the ability to maintain stable flight, after changing its formation by rotating motion of each arm individually, that has been presented in \cite{ETHmorphing}. Another foldable quadrotor design was presented in \cite{AgileRoboticFliers}, where the platform was able to decrease its wide-span, by changing the orientation of its propellers based on an actuated elastic mechanism. Furthermore, in \cite{FOLLY2} a self-foldable quadrotor has been presented with a gear-based mechanism to control the contraction and expansion of the four arms simultaneously. This approach allowed for two possible configurations namely either fully expanded when the drone is deployed or fully contracted when the drone is on the ground. A passive foldable quadrotor has been presented in \cite{passiveMorphing} that utilizes springs for altering its formation. The maneuverability of this design, while the drone is in its reduced form, was limited and the quadrotor can traverse only for a short time through narrow gaps. Finally, a sliding arm quadrotor has been presented in \cite{SlidingArmQuadcopter} from a modeling and control point of view.

Besides transformable quadrotor platforms, other novel aerial vehicles that can alter their structure have been presented in the last years.  \textit{DRAGON} is a dual rotor multilink aerial robot that alters its formation with the use of multiple servos while flying \cite{dragon} able to traverse through gaps. In the area of aerial grasping the robotic platform in \cite{AerialGrasping}, consists of multiple links that can be actuated to adopt its overall shape for the handling of large objects. The ability of robots to adapt to their environment and on the needs of a mission, it has been investigated in \cite{TheShapeshifter}. Where the concept of a hybrid platform has been presented with a combination of multiple robots which can collaboratively fly and roll in different formations.

\subsection{Contribution}
The novelty of this work stems from the design of a switching \gls{mpc} to support the online structural reformation of a foldable quadrotor. The novel proposed control framework can count for state and control signal constraints during the shape transition and adapt to the induced model variations due to the shape transformation. The selected control scheme is evaluated under iterative simulations during navigation of various paths, while the platform is executing sequential transformations. Finally, the effect of the platform's shape during motion is investigated.
\subsection{Outline}
The rest of the article is structured as follows. Section \ref{sec:pf} discusses the modelling of the foldable quadrotor and the control problem. Section \ref{sec:control} discusses the design of the \gls{mpc} attitude controller while Section \ref{sec:sim} presents the simulation results. Finally, concluding remarks are given in Section \ref{sec:con}.
\section{Modeling} \label{sec:pf}

\begin{figure}[htb]
\begin{center}
\includegraphics[width=\columnwidth]{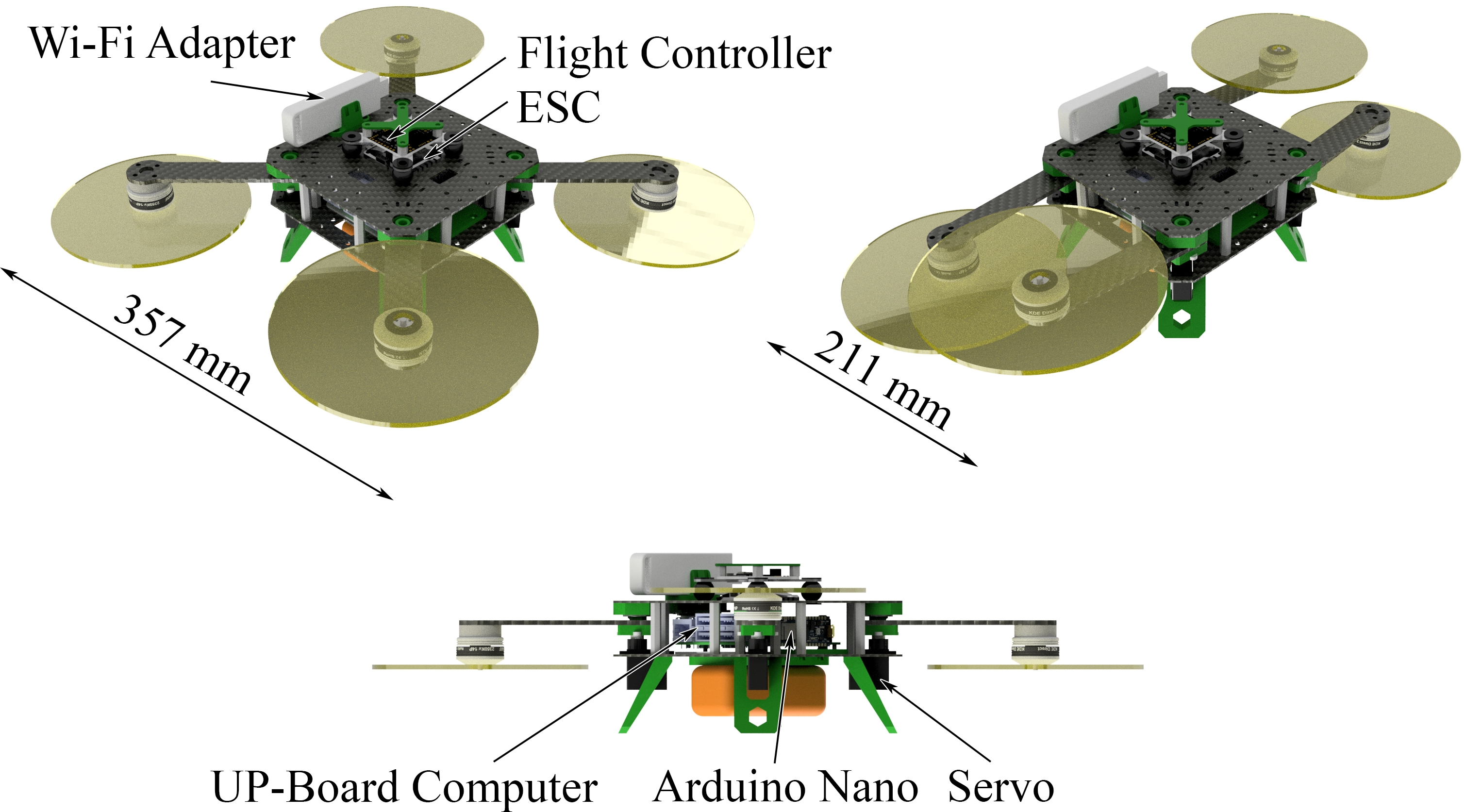}
\caption{Isometric view of the foldable drone conceptual design in X and H morphology and a side view displaying the various components.}
\label{fig:conceptual}
\end{center}
\end{figure}

\begin{figure}[htb]
\begin{center}
\includegraphics[width=0.8\columnwidth]{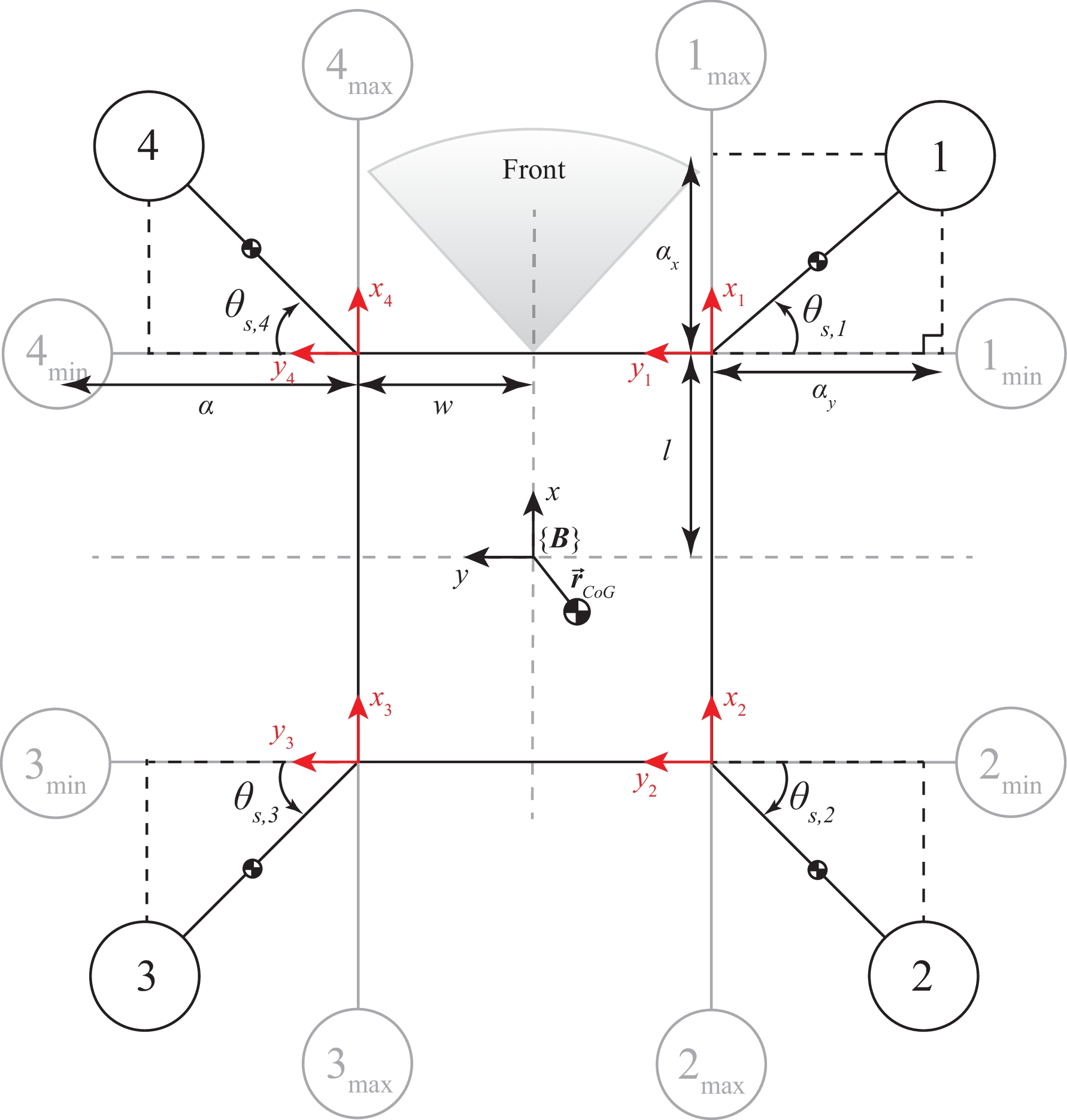}
\caption{2D representation of the foldable quadrotor with highlighted the main geometrical properties}
\label{fig:2Dgeom}
\end{center}
\end{figure}

In Fig.~\ref{fig:conceptual}, the conceptual design of the foldable quadrotor is depicted in isometric view for different formations. As it is indicated, the arms of the quadrotor are connected on servos thus they can rotate around the $z$-axis. To overcome possible collision between the propellers at the extreme angles, the motors have been placed alternately upside down. It is important to mention that any changes on the formation of the arms coming directly from rotation around the $z$-axis of the \gls{mav}. Thus, the geometry varies only related to the $x$ and $y$-axis resulting into a planar-varying geometry as it is depicted in Fig.~\ref{fig:2Dgeom}. 

In this work both symmetrical and asymmetrical formations are studied. The major changes impacting the quadrotor from the different formations are coming from \gls{cog} variations  and changes in \gls{moi} matrix. The foldable quadrotor is a 6-\gls{dof} object with a Body-Fixed Frame $\textbf{B}$ located at the \gls{gc} of the vehicle. The arms of the platform are able to perform only 1-\gls{dof} motion around the $z$-axis.

Platform's \gls{cog} is located at $\pmb{r}_{\mathrm{CoG}}\in\mathbb{R}^3$ distance from the \gls{gc}. The offset vector $\pmb{r}_{\mathrm{CoG}}$ is calculated by taking into account every component of the quadrotor \gls{cog} position vector $\pmb{r}_{(.)}\in\mathbb{R}^3$. In this article we consider the following dominant components that characterize the geometry of the \gls{mav}. These are the main body of the quadrotor with mass $m_b$ and dimensions $(2l \times 2w \times 2h)$ denoting the length, width and height respectively. The arms located at the four corners of the \gls{mav} with mass $m_{a,i}$ and offset $\pmb{r}_{a,i}$, where $i\in\mathbb{Z}[1,4]$ is the identification number of the individual components. Finally, the combination of the motor, rotor and propeller is considered one component with mass $m_c$ and offset $\pmb{r}_{c,i}$ from the \gls{gc} and its own \gls{cog}.

In Fig~\ref{fig:2Dgeom}, $\theta_{s,i}\in\mathbb{S}^{1}$ are the angles of the servos actuating the arms, while the offset $\pmb{r}_{CoG}$ between the \gls{gc} of the vehicle is:
\begin{equation}\label{eq:CoG1}
 \pmb{r}_{CoG} = \frac{m_{b}\pmb{r}_{b} + \sum_{i=1}^{4} (m_{a}\pmb{r}_{a,i} + m_{c}\pmb{r}_{c,i}) }{m_{b} + \sum_{i=1}^{4} (m_{a}+ m_{c})}
\end{equation}
Altering the formation of the platform, by actuating the servos, the total mass of the platform remain the same or:
\begin{equation}\label{eq:total_mass}
 m = m_{b} + \sum_{i=1}^{4} (m_{a}+ m_{c} ),
\end{equation} 
since the distance of the components' \gls{cog} is a function of the servos angle $\theta_{s,i}$ the \eqref{eq:CoG1} can be written as,
\begin{equation}\label{eq:CoG2}
 \small{\pmb{r}_{CoG} = \frac{1}{m}\left(
 m_{b}\pmb{r}_{b} + 
 \sum_{i=1}^{4} (m_{a}\pmb{r}_{a,i}(\theta_{s,i}) + m_{c,i}\pmb{r}(\theta_{s,i}))\right)}. 
\end{equation}
The distance, from the geometric center to the servo, is constant and equal to the dimensions of the body $[w,~l]^\top$. The offset vectors $\pmb{r}_{(.)}$ can be calculated either online or offline with the knowledge of the angle $\theta_{s,i}$. The offset on the $z$-axis of every component to the \gls{gc} is constant, since it does not change when they rotate around $z$-axis, however, the offsets $\pmb{r}_{.,x}$ and $\pmb{r}_{.,y}$ need to be recalculated. The angle $\theta_s$ of the servo is assumed to be known (Fig.~\ref{fig:2Dgeom}). For this study, the position $(.)_{\min}$ in Fig.~\ref{fig:2Dgeom} denotes the minimum angle that the arm can rotate \ang{0}, while the $(.)_{\max}$ denotes the maximum angle \ang{90}.

Under the assumption that the \gls{cog} of the combination motor, rotor and propeller located in their center, the euclidean distance of the \gls{cog} from the servo is $\alpha$. Based on the geometrical properties the new position, where the thrust is generated for every arm can be calculated by:
\begin{subequations}
\label{eq:test2}
\begin{align}
&\footnotesize{\pmb{r}_{c,1} =  \left[\arraycolsep=2.2pt\def\arraystretch{1.3}\begin{array}{ccc} - w - \alpha \cos{\theta_{s,1}} - r_{\text{CoG},y}, & l + \alpha \sin{\theta_{s,1}} - r_{\text{CoG},x}, & z \end{array}\right]^\top} \\ 
&\footnotesize{\pmb{r}_{c,2} =  \left[\arraycolsep=2.2pt\def\arraystretch{1.3}\begin{array}{ccc}
- w - \alpha \cos{\theta_{s,2}} - r_{\text{CoG},y}, & - l - \alpha \sin{\theta_{s,2}} - r_{\text{CoG},x}, & z  \end{array}\right]^\top} \\ 
&\footnotesize{\pmb{r}_{c,3} =  \left[\arraycolsep=2.2pt\def\arraystretch{1.3}\begin{array}{ccc} w + \alpha \cos{\theta_{s,3}} - r_{\text{CoG},y}, & - l - \alpha \sin{\theta_{s,3}} - r_{\text{CoG},x}, & z  \end{array}\right]^\top} \\ 
&\footnotesize{\pmb{r}_{c,4} =  \left[\arraycolsep=2.2pt\def\arraystretch{1.3}\begin{array}{ccc} w + \alpha \cos{\theta_{s,4}} - r_{\text{CoG},y}, &  l + \alpha \sin{\theta_{s,4}} - r_{\text{CoG},x}, & z  \end{array}\right]^\top}
\end{align}
\end{subequations}
The varying positions of the arms result into a varying control allocation, since it affects the position of the motors and directly the torques around the $x,y$-axis.
\begin{equation}\label{eq:controlmix}
\small{\begin{bmatrix}
T\\
\tau_x \\
\tau_y \\
\tau_z 
\end{bmatrix}}
=\small{\begin{bmatrix}
b & b & b & b\\
 r_{c,1_x} & r_{c,2_x} & r_{c,3_x} & r_{c,4_x} \\
 r_{c,1_y} & r_{c,2_y} & r_{c,3_y} & r_{c,4_y} \\
-\kappa & \kappa & -\kappa & \kappa\\
\end{bmatrix}}
\small{\begin{bmatrix}
f_1 \\ f_2 \\ f_3 \\ f_4
\end{bmatrix}}
\end{equation}
 where $r_{c,i_x}$ and $r_{c,i_y}$ denote the first and second element respectively of the $\pmb{r}_{c,i}$. Finally, $b$ and $\kappa$ are coefficients related to the thrust and torque  respectively. 




In addition, the \gls{moi} of the platform varies as the arms change position. Each component is characterized by its own \gls{moi} matrix. 
The total \gls{moi} of the platform can be calculated from the individual \gls{moi} with the use of the parallel axis theorem at the body frame $\textbf{B}$.
\begin{equation} \label{eq:total_intertiaB}
\small{(\pmb{I}_{mav})_{\textbf{B}} = (\pmb{I}_{b})_{\textbf{B}} + \sum (\pmb{I}_{a_i})_{\textbf{B}} + \sum (\pmb{I}_{m_c})_{\textbf{B}} }
\end{equation}
For this article, as distinct formations are considered for the switching \gls{mpc}, the \gls{moi} have been extracted directly from the CAD model for all the different configurations, instead of computing them from the geometrical properties.
\section{Control Structure} \label{sec:control}

The purpose of the attitude switching \gls{mpc} is to track the desired angles $\phi_d, \theta_d$ and $\psi_d$ given from a higher level trajectory controller. 
For the trajectory tracking, a high level \gls{mpc} is formulated and used based on the linearized translation euler model~\cite{Kamel2017}.  Thus, the following states are considered for the switching \gls{mpc},
\begin{equation} \label{eq:LLstates}
\pmb{x}_{smpc} = \left[\begin{array}{ccccccccc} \phi & \theta & \psi & \dot{\phi} & \dot{\theta} & \dot{\psi} & \tau_x & \tau_y & \tau_z \end{array}\right]^\top
\end{equation}
The angular acceleration, rate and torques are given from the Newton-Euler law as:
\begin{equation} \label{eq:newton-euler}
 \dot{\pmb{\omega}} = \pmb{I}^{-1}(-\pmb{\omega}\times \pmb{I} \pmb{\omega} + \pmb{\tau}),
\end{equation}

\noindent where we consider the inertia matrix with zero off-diagonal elements:
\begin{equation} \label{eq:intertiax}
 (\pmb{I}_{mav})_{B} = \pmb{I} = \left[\begin{array}{ccc}
 I_{xx} & I_{yy}& I_{zz}
\end{array}\right] \pmb{I}_{3}
\end{equation}

\noindent At this point, it should be noted that $\omega \neq \dot{\eta}$, where ${\eta} =[\phi~\theta~ \psi ]^\top $. The transformation matrix for the angular velocities, from the inertial frame to the body frame, is $\pmb{W_{\eta}}$.
 

 $$\pmb{\omega}=\pmb{W_{\eta}}\dot{\pmb{\eta}},~~~
 \left[\begin{array}{c} {\omega_x} \\ {\omega_y} \\ {\omega_z} \end{array}\right] = 
\left[\begin{array}{ccc} 1 & 0 & -s\theta \\ 0 & c\phi & c\theta s\phi \\ 0 & -s\phi & c\theta c\phi \end{array}\right] 
\left[\begin{array}{c} \dot{\phi} \\ \dot{\theta} \\ \dot{\psi} \end{array}\right]
$$
\begin{equation} \label{eq:newton-eulerBody}
\dot{\pmb{\omega}} = \pmb{I}^{-1}(-\pmb{W}\dot{\eta}\times \pmb{I} \pmb{W}\dot{\pmb{\eta}} + \pmb{\tau}) 
\end{equation}

As far as the torques' dynamics are concerned, following a similar approach to \cite{ETHmorphing}, they have been assumed to follow the dynamics of a first order system.
\begin{equation} \label{eq:taudynamics}
\dot{\tau} = \frac{1}{\tau_{\alpha}}(\tau_{d}-\tau)
\end{equation}

\noindent where $\tau_{\alpha}$ is the time constant. After linearizing \eqref{eq:newton-eulerBody} at $\omega=0$ and $\tau=0$ it results into the following linear system. 

\begin{equation}
\begin{bmatrix}
\pmb{\omega}\\
\dot{\pmb{\omega}} \\
\dot{\pmb{\tau}} 
\end{bmatrix}
=\begin{bmatrix}
\pmb{0} & \pmb{I}_3 & \pmb{0} \\
\pmb{0} & \pmb{0} & \pmb{I}^{-1} \\
\pmb{0} & \pmb{0} & -\frac{1}{\tau_{\alpha}}\pmb{I}_3 \\
\end{bmatrix}
\begin{bmatrix}
\pmb{\eta} \\ \pmb{\omega} \\ \pmb{\tau}
\end{bmatrix} 
+
\begin{bmatrix}
\pmb{0} \\ \pmb{0} \\ \frac{1}{\tau_{\alpha}}\pmb{I}_3
\end{bmatrix} 
\end{equation}

\begin{figure*}[htb]
\begin{center}
\includegraphics[width=0.8\textwidth]{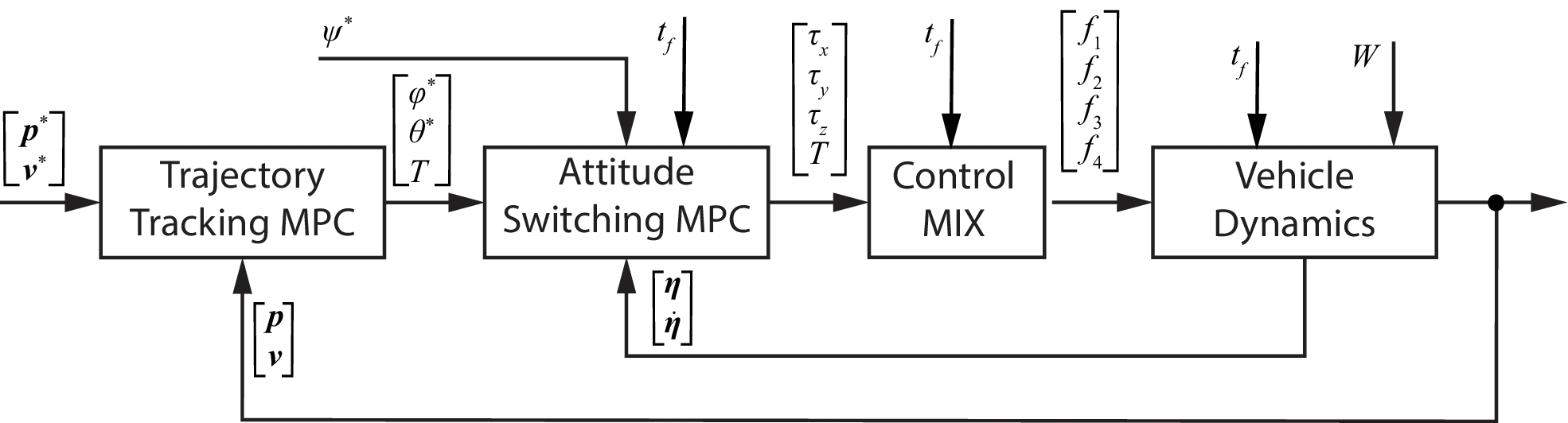}
\caption{Block diagram displaying the overall control scheme proposed for controlling the foldable quadrotor.}
\label{fig:blockdiagram}
\end{center}
\end{figure*}
 


\subsection{Linear Parameter Varying System}
 
The attitude modeling of the quadrotor from the linearization stage depends on the inertia matrix. While the inertia matrix varies based on the formation of the platform the system is subject to parametric changes. The resulting \gls{lpv} system has the following form,
\begin{equation}
 \pmb{x}_{k+1} = \pmb{A}(\theta_{S,k})\pmb{x}(k) + \pmb{B}(\theta_{S,k})\pmb{u}(k)
\end{equation}
where $\pmb{x}(k) \in \mathcal{R}^{n \times 1}$ denotes the system states and $\pmb{u}(k)\in \mathcal{R}^{m \times 1}$ is the input vector.

For the case of the linear switching \gls{mpc} the following optimization is considered:
\begin{subequations}
\label{eq:mpc}
\begin{align}
& \text{minimize} && 
\sum\limits_{k=0}^{N-1}\left( \Delta \pmb{x}_{k}^\top \pmb{Q}_x \Delta \pmb{x}_{k} + \pmb{u}_{k}^{T} \pmb{R}_u \pmb{u}_{k} \right)\\
& \text{subject to}
& & \pmb{x}_{k,\text{min}} \leq \pmb{x}_{k} \leq \pmb{x}_{k,\text{max}}, \; k=1,...,N_p, \\
&&& \Delta \pmb{u}_{\text{min}} \leq \Delta \pmb{u}_k \leq \Delta \pmb{u}_{\text{max}}, \; k=1,...,N_p, \\
&&& \pmb{x}_{k+1|k}=f(\pmb{x}_{k|k},\pmb{u}_{k|k},\pmb{\theta_{s}}_{k|k}), \; k \geq 0, \\
&&& \pmb{x}_0 =\pmb{x}(t_0)
\end{align}
\end{subequations}
%
 %
\noindent where $\Delta \pmb{x}_{k} = \pmb{x}_{k}^{*} - \pmb{x}_{k}$ and $\Delta \pmb{u}_{k} = \pmb{u}_{k|k} - \pmb{u}_{k|k-1}$, while $N_p,N_c$ denote the prediction and control horizon respectively. $\pmb{Q}_x \succeq 0$ and $\pmb{R}_u \succ 0$ are the penalty on the state error and on the control input respectively, while the bounds of the constraints are denoted as $(.)_{\text{min},\text{max}}$. The state update of the optimization problem is a function of the current states, inputs and the angle of the arms $\pmb{\theta}_{s}$ and $\pmb{x}_0$ are the initial state conditions.

The complete control scheme is displayed in Fig.~\ref{fig:blockdiagram}. For a reference profile of positions $\pmb{p}^* =[x,y,z]^\top$ and velocities $\pmb{v}^* =[\dot{x},\dot{y},\dot{z}]^\top$ the trajectory tracking \gls{mpc} generates roll, pitch and thrust commands. Next the attitude switching \gls{mpc} selects the appropriate model based on the switching variable $t_f$ which indicates the formation of the platform (X-H-Y and T). The computed torques and thrusts from the switching \gls{mpc} are given to the parametric varying control mix as defined in \eqref{eq:controlmix}, which results the necessary forces for the motors.
\section{Simulation Results} \label{sec:sim}
To evaluate the performance of the attitude switching \gls{mpc}, two indicative types of simulations are presented. The first scenario assumes that the quadrotor takes off and hovers at a specific height, while cycling through the four different formations denoted as $\text{X},~\text{H},~\text{Y},~\text{T}$, while the states of the plant are subject to additive noise. The second scenario tests the ability of the foldable quadrotor to follow a square trajectory while cycling again through the different configurations. 

The utilized parameters for the nonlinear quadrotor are: a total mass of 1kg and arm length 0.15m. To increase the accuracy, the inertia tensors are computed from the CAD model directly for the different formations and they are given in the following table. 

{\renewcommand{\arraystretch}{1.3}
\begin{table}[htb]
 \centering
 \caption{Inertia Tensor Values for the different formations}
 \label{tab:in}
\begin{tabular}{|c|c|c|c|}
 \hline
 & $I_{xx}$ & $I_{yy}$ & $I_{zz}$\\ 
 \hline
 X & 0.004233 & 0.004380 & 0.007834\\ 
 \hline
 H & 0.005885 & 0.001812 & 0.006918 \\ 
 \hline
 Y & 0.005042 & 0.003096 & 0.007369\\ 
 \hline
 T & 0.003654 & 0.003917 & 0.006792 \\ 
 \hline
\end{tabular}
\end{table}
}

The attitude switching \gls{mpc} has a prediction horizon of $N_{p}=40$ and a control horizon of $N_{c}=12$ with a sampling time of 0.01sec. The input weights are set to $\pmb{R}_{u}=\text{diag}(80, 80, 120)$, while the yaw reference is kept at $\psi^*=0$. The states weight matrix is $\pmb{Q}_{x}=\text{diag}(40,40,40,80,80,80,0.1,0.1,0.1)$, while the rate constraints are $\pmb{\Delta_{u}}\leq |[0.03, 0.03, 0.03]^\top|$ and the input constraints $\pmb{u}\leq |[0.1, 0.1, 0.1]^\top|$. The weights, penalties and constraints are identical for both simulations. Since on this preliminary evaluation, the interest is focused on maintaining a stable flight, the formation change of the platform set to happen at specific time-instances. 
\subsection{Attitude Control Simulation}
\begin{figure}[htb]
\begin{center}
\includegraphics[width=\columnwidth]{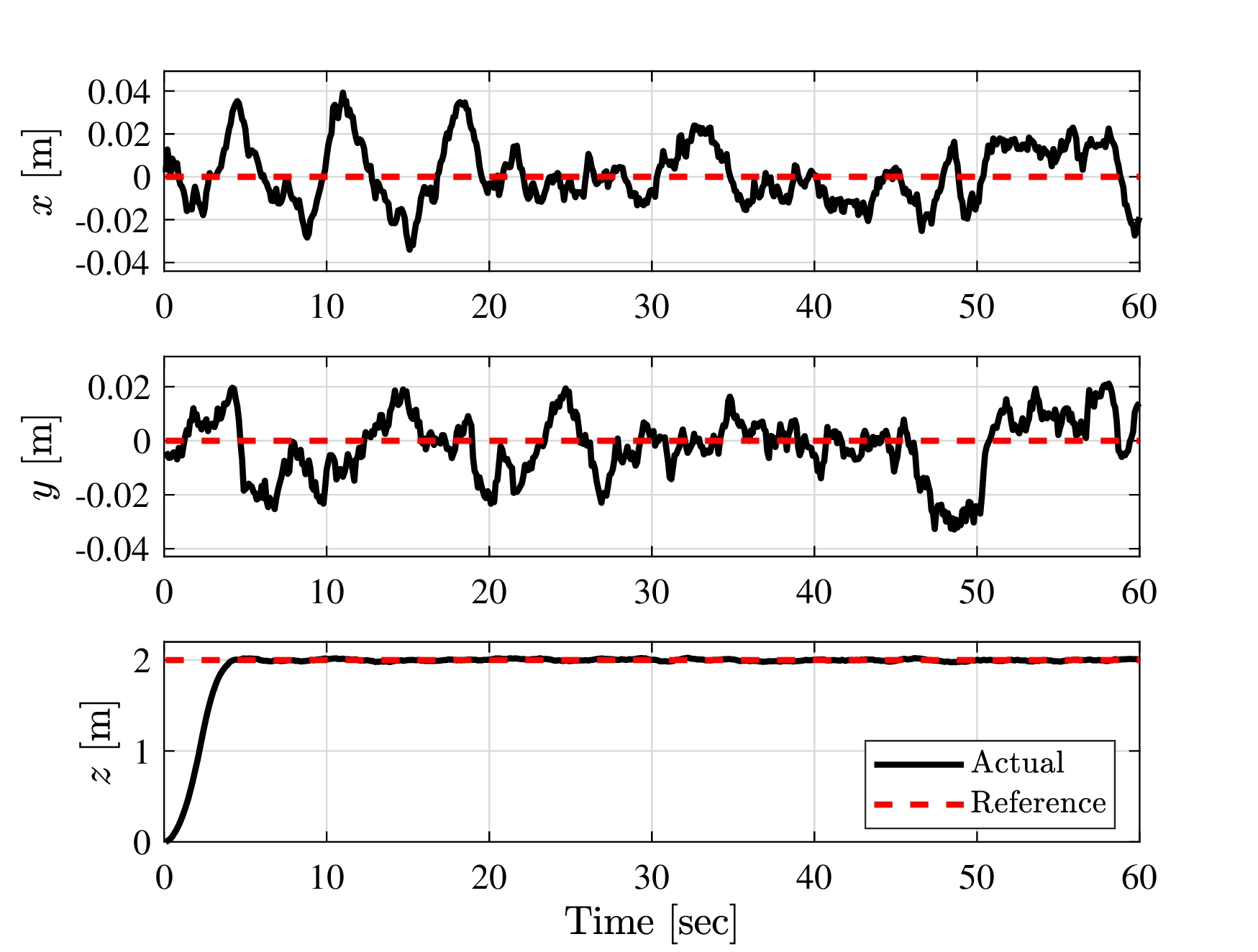}
\caption{Time response of the foldable quadrotor position in $x,y$ and $z$-axis during position hold, while changing its formation.}
\label{fig:3dhover}
\end{center}
\end{figure}

For the first simulation the reference signal hold position at [0,0,2] meters in $x,y$ and $z$ axis respectively, while a switching signal $t_{f}$ is sent every 15 seconds to update the formation of the drone, thus forcing the controller as well to update its model. Fig.~\ref{fig:3dhover} shows the time response performance of the position vector. It can be noticed that the error remained under 0.04m for $x$ and $y$, while for $z$ under 0.02m after it reached the steady state. 

%

\begin{figure}[htb]
\begin{center}
\includegraphics[width=\columnwidth]{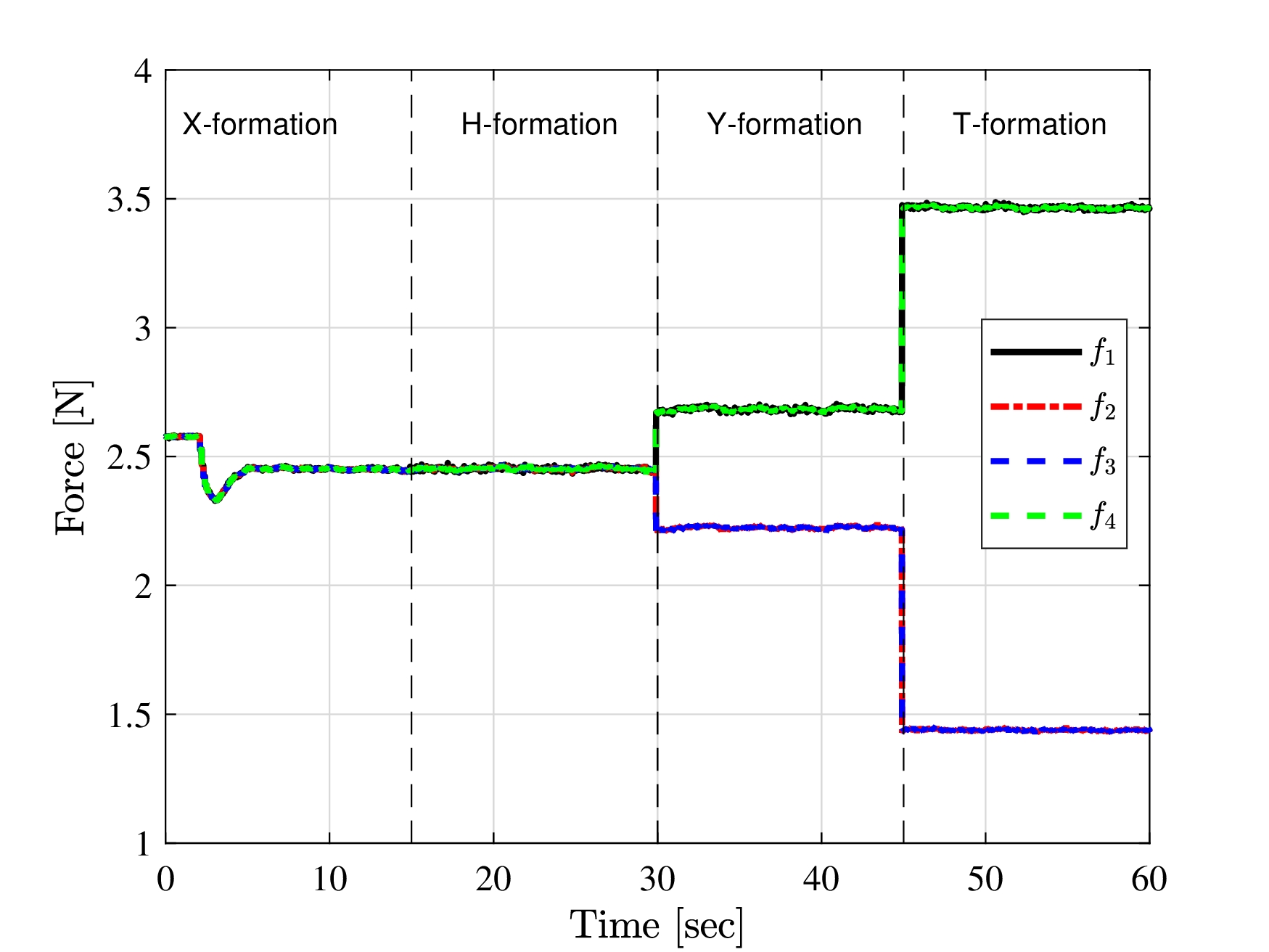}
\caption{Motor commands during position hold, while altering the morphology of the platform.}
\label{fig:mc_hover}
\end{center}
\end{figure}

As far as the performance of the attitude controller is concerned, Fig.~\ref{fig:mc_hover} shows the generated force levels required for each motor to achieve position hold. It can be noticed that while the platform is in H or X configuration, which are both symmetrical formations, there is no major impact from the configuration change. On the other hand, the formations Y and T, which result to a major change in the geometry of the platform and variation of the \gls{cog} that has a great impact on the motor forces. For the T formation, the motor$_{1,4}$ need to generate approximately 3.5N, while the motor$_{2,3}$ about 1.4N with the total force to be equal to the gravity force $g$ as happens in all other configurations. 

\begin{figure}[htb]
\begin{center}
\includegraphics[width=\columnwidth]{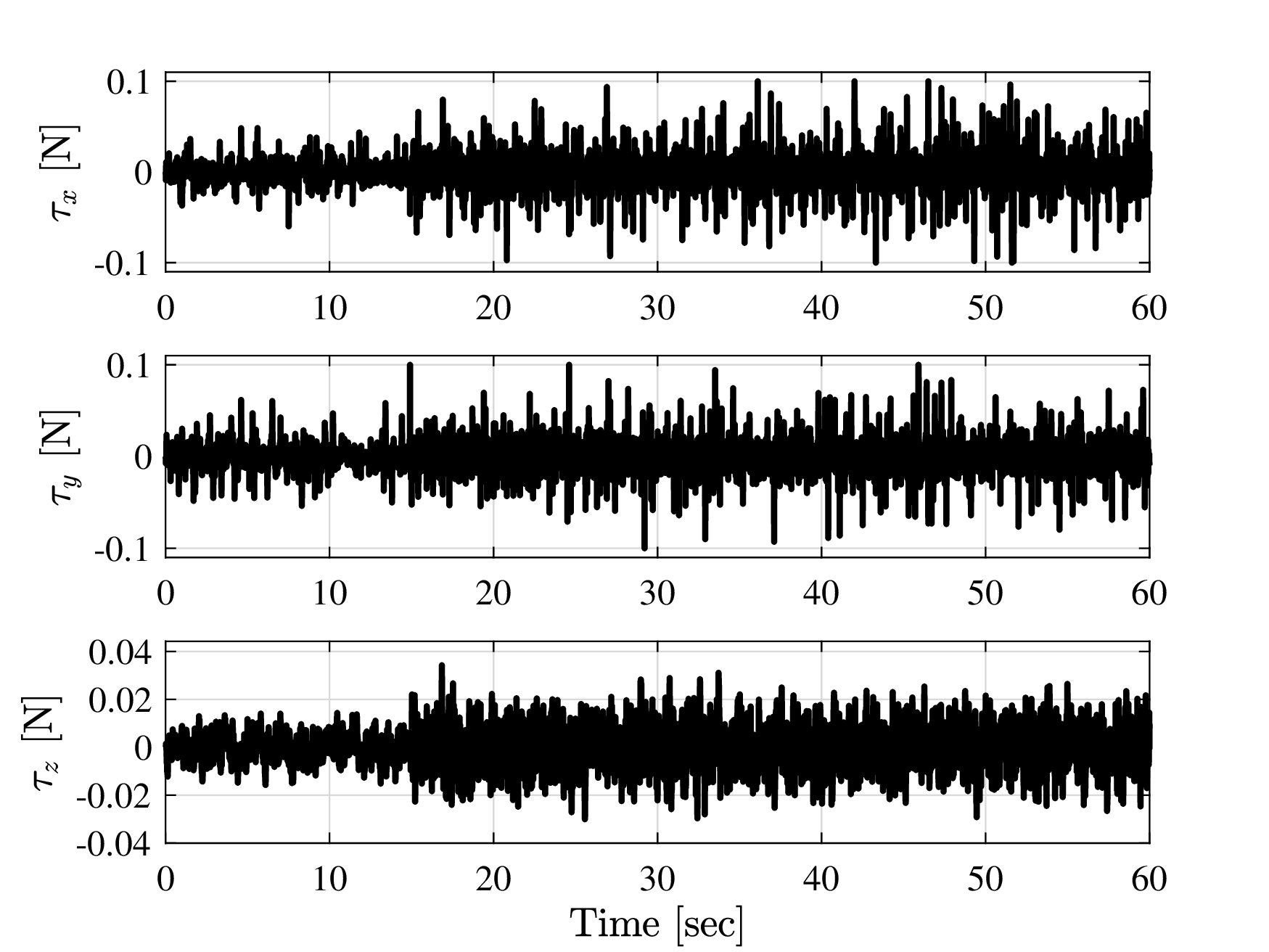}
\caption{Torque output response of the switching \gls{mpc} during position hold simulation.}
\label{fig:torque_hover}
\end{center}
\end{figure}

In Fig.~\ref{fig:torque_hover}, the torque input generated from the switching \gls{mpc} is illustrated. The presented time response of the torques reaches close to the boundaries in an effort to maintain the position of the platform. Despite the high amplitude noise the controller successfully maintains the position without violating the pre-defined constraints. 

\subsection{Trajectory Tracking Simulation}

The trajectory tracking of the linear \gls{mpc} has a sampling time of 0.1sec. The input weights are set to $\pmb{R}_{u}=\text{diag}(25, 25, 8)$ for the roll pitch and thrust, while the yaw reference is kept at $\psi^*=0$. The states weights are set to $\pmb{Q}_{u}=\text{diag}(40,40,60,80,80,80,0.1,0.1)$, while the states are defined as $[\pmb{p},\pmb{v},\phi^*,\theta^*]^\top$. The rate constraints are $\pmb{\Delta_{u}}\leq |[0.3, 0.3, 0.0025]^\top|$ and the input constraints $ \pmb{u}\leq |[12, 12,\infty]^\top|$.

\begin{figure}[htb]
\begin{center}
\includegraphics[width=\columnwidth]{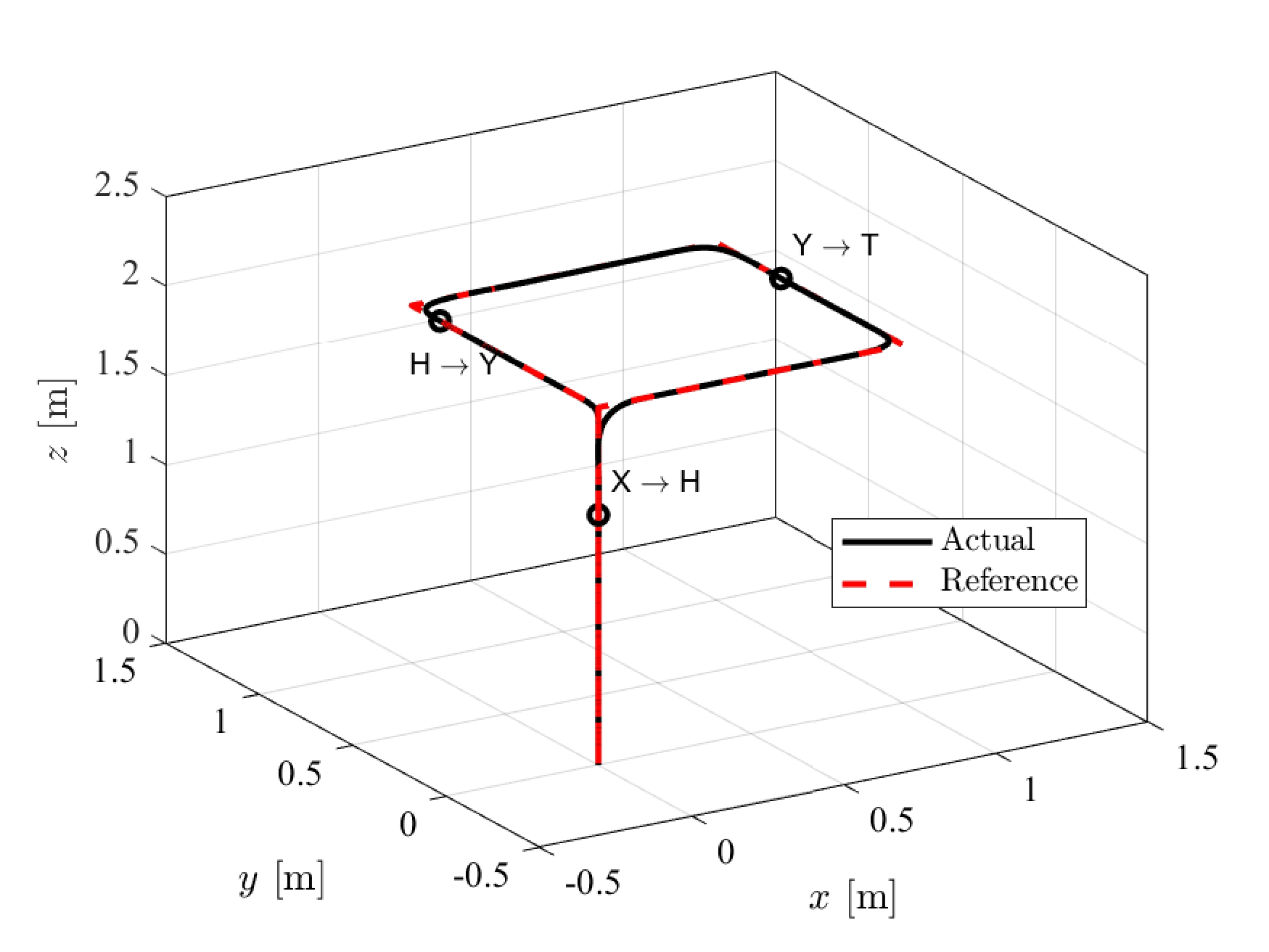}
\caption{Foldable drone trajectory tracking performance for a given square path.}
\label{fig:3dsquare}
\end{center}
\end{figure}

The tracking performance of the foldable quadrotor is illustrated in Fig.~\ref{fig:3dsquare}. The \gls{mav} successfully tracks all the way-points despite the formation changes even when the reformation occurs close to a turn. 

\begin{figure}[htb]
\begin{center}
\includegraphics[width=\columnwidth]{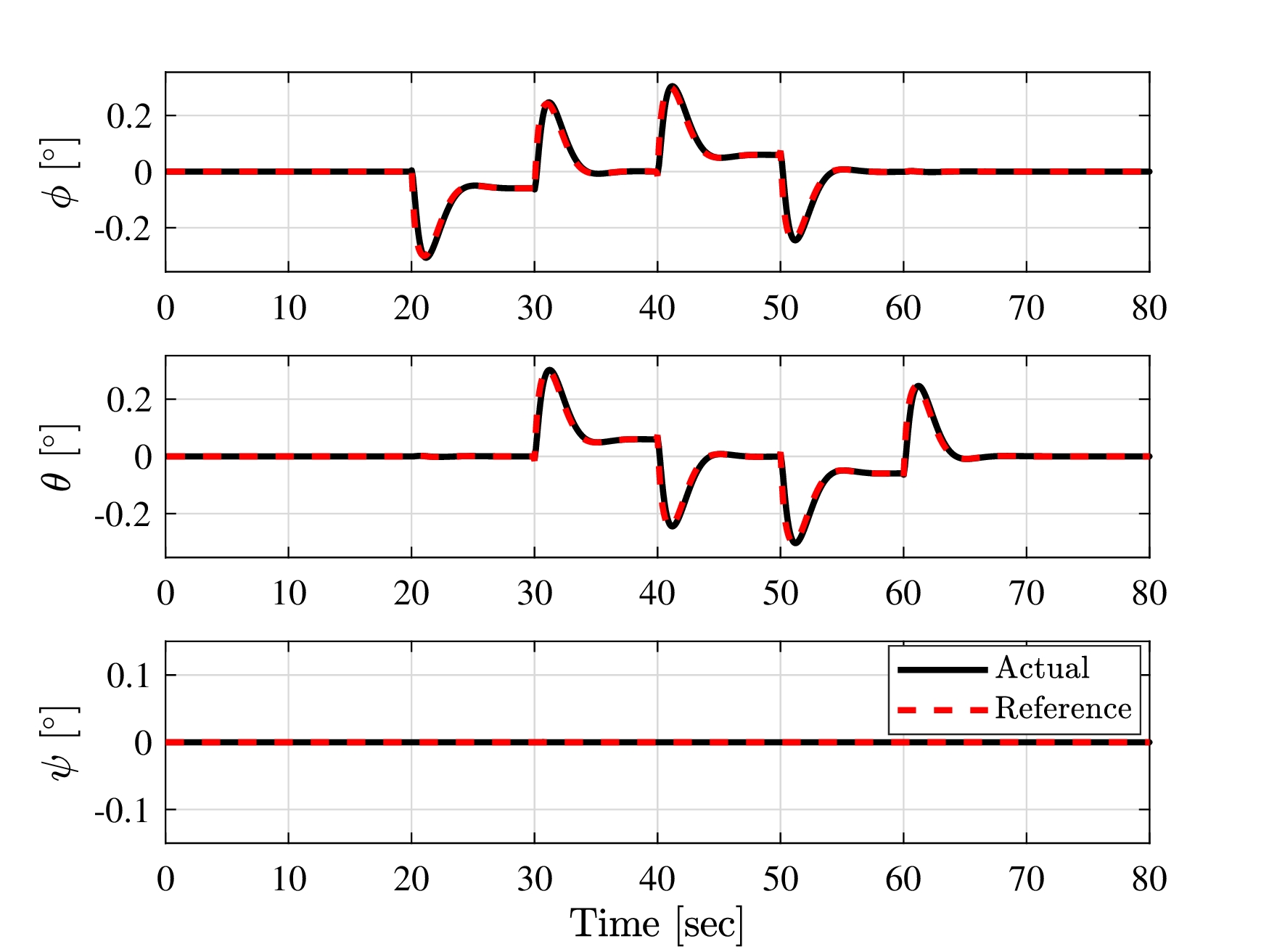}
\caption{Roll, pitch and yaw references and responses during the trajectory tracking simulation of the switching attitude \gls{mpc}.}
\label{fig:angle_square}
\end{center}
\end{figure}

\begin{figure}[htb]
\begin{center}
\includegraphics[width=\columnwidth]{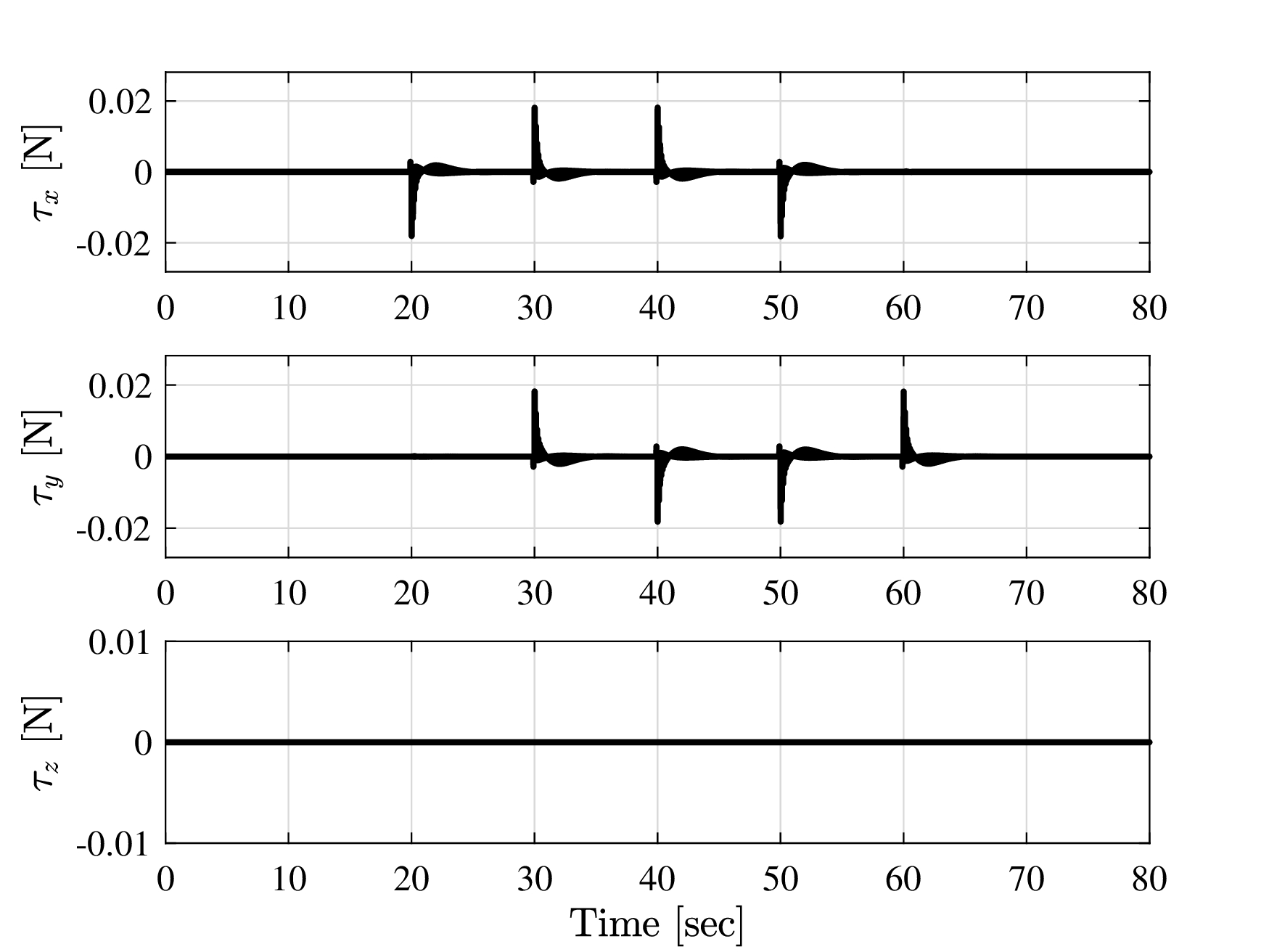}
\caption{Torque output response of the switching \gls{mpc} during the trajectory tracking simulation.}
\label{fig:torque_square}
\end{center}
\end{figure}

The reference angles $\theta^*$ and $\phi^*$ generated from the trajectory controller are given in \ref{fig:angle_square} (red dashed line), while the yaw is forced to $\psi^*=0$, are successfully tracked from the switching \gls{mpc} (black line). Fig.~\ref{fig:torque_square} shows the switching controller torque outputs. As expected, the $\tau_z$ remained zero, while the torques around the $x$ and $y$ axis remained under 0.02N, resulting into a smooth transition throughout the entire trajectory. 

\begin{figure}[htb]
\begin{center}
\includegraphics[width=\columnwidth]{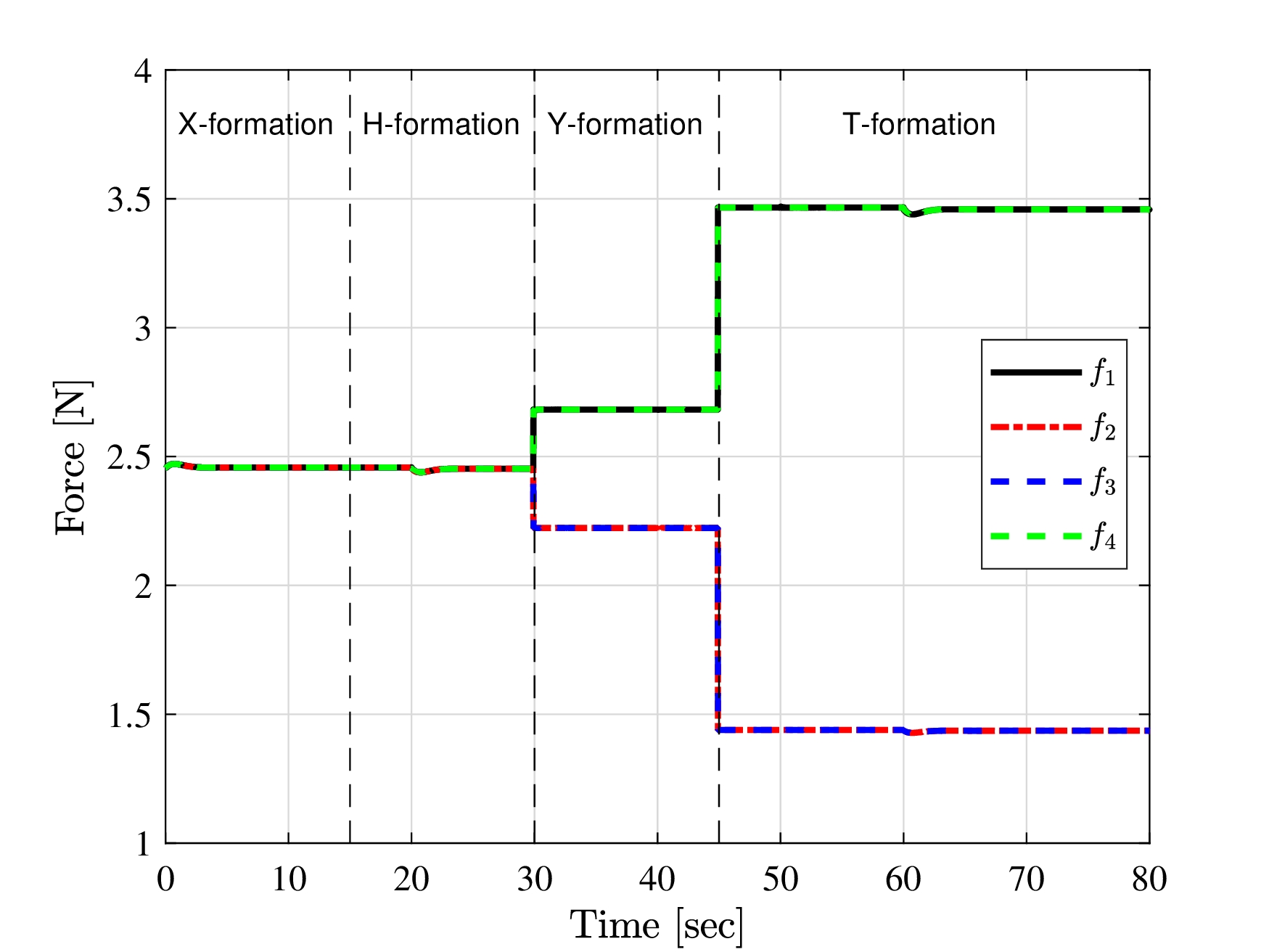}
\caption{Motor commands during trajectory tracking, while altering the morphology of the platform.}
\label{fig:mc_square}
\end{center}
\end{figure}

Finally, similarly to the position hold simulation, Fig \ref{fig:mc_square} the morphology of the platform has a major impact on the required force by each motor. 

\section{Conclusions} \label{sec:con}
 This article presented a switching \gls{mpc} for the online structural reformation of a foldable quadrotor. To evaluate the efficacy of the control scheme, simulation trials have been performed during online reformations. The switching \gls{mpc} scheme has presented the ability to successfully maintain the performance, despite the alternating of configurations resulting into a stable flight during all the simulation trials. The overall position error remained under 4cm in $x,y$ and $z$-axis. The incorporation of the switching \gls{mpc}, with a trajectory tracking controller, was successful as the first was able to regulate properly and track the desired angles with a minimum error. The performance of the switching controller is characterized by accurate tracking of the way-points for the case of following a square trajectory, while changing its formation. During the simulations, the error remained in the level of centimeters. Future work will tackle the challenge of deploying the switching \gls{mpc} in extended experimental evaluations. 
\bibliographystyle{IEEEtran}
\bibliography{References}
\end{document}